\newcommand{\ig}[1]{}

\documentclass{llncs}
\usepackage{graphicx}
\usepackage{named}
\usepackage{amssymb,amsmath}

\begin{document}
\title{A Partial Taxonomy of\\
Substitutability and Interchangeability}
\author{Shant Karakashian${}^1$, Robert Woodward${}^1$, Berthe Y. Choueiry${}^1$,
Steven D. Prestwich${}^2$ and Eugene C. Freuder${}^2$}
\institute{
${}^1$ Constraint Systems Laboratory,
University of Nebraska-Lincoln, USA
{\tt \{shantk,rwoodwar,choueiry\}@cse.unl.edu}\\
\vspace{0.06in}
${}^2$ Cork Constraint Computation Centre,
Department of Computer Science, University College Cork, Ireland
{\tt \{s.prestwich,e.freuder\}@4c.ucc.ie}}
\maketitle
\begin{abstract}
Substitutability, interchangeability and related concepts in
Constraint Programming were introduced approximately twenty years ago
and have given rise to considerable subsequent research.  We survey
this work, classify, and relate the different concepts, and indicate
directions for future work, in particular with respect to making
connections with research into symmetry breaking.  This paper is
a condensed version of a larger work in progress.
\end{abstract}

\section{Introduction}

Many important problems in computer science, engineering and
management can be formulated as Constraint Satisfaction Problems
(CSPs). A CSP is a triple $(V,D,C)$ where $V$ is a set of variables,
$D$ the set of their domain values, and $C$ a set of constraints on
the variables that specify the permitted or forbidden combinations of
value assignment to variables. A solution to a CSP is an assignment of
values to all variables such that all constraints are satisfied. CSPs
are usually solved by interleaving backtrack search with some form of
constraint propagation, for example forward checking or arc
consistency.

Constraint problems often exhibit symmetries.  A great deal of
research has been devoted to {\it symmetry breaking\/} techniques in
order to reduce the size of the search space
\cite{Symcon}.  The earliest works on symmetry breaking include
\cite{Glaisher:1874,Brown88}.  In this paper we will not survey the
large literature on symmetry breaking, but a recent survey can be
found in \cite{gent2006symmetry}.

Interchangeability, proposed in a seminal paper by Freuder
\shortcite{freuder1991eliminating}, is one of the first forms of
symmetry identified for CSPs.  Importantly, it is also the first
method proposed for {\it detecting\/} symmetry as opposed to having a
constraint programmer manually specify it.  Although there has been
since then a steady flow of research papers developing this concept in
both theory and practice, it has been relatively neglected compared to
other forms of symmetry.  This situation is surprising: While in its
basic form, interchangeability is a special case of value symmetry,
its various extensions (already proposed in the 1991 paper) make it a
more general concept than is sometimes perceived, and anticipate some
subsequent developments in symmetry definition and breaking.  The
comparison with the various types and definitions of symmetry
\cite{benhamou1994study,cohen2006symmetry} will be discussed in the
longer version of this paper.

The goal of our endeavor is to analyze the research conducted so far
on interchangeability, relate it to symmetry, and identify
opportunities for future research.  This paper is a work in progress
and a first step towards our goal.  Our survey is partial and far from
complete and we welcome the feedback of the readers and workshop
participants.

The advantages of detecting and exploiting interchangeability have
been established on random problems, benchmarks, and real-world
applications.  In backtrack search, the advantages are mainly the
reduction of the search space and the search effort
\footnote{Most importantly, by factoring out no-goods \cite{Choueiry:sara02}.}, 
and the attainment of multiple solutions by bundling.  In local
search, interchangeability is used to locally repair partial solutions
\cite{petcu2004applying}.  Real-world applications include nurse
scheduling \cite{Weil:95} and resource allocation in hospitals
\cite{choueiry-ijcai1995abstraction}.

This paper is structured as follows.  In Section~\ref{concepts}, we
give the definitions of the basic interchangeability concepts and
relate them to each other.  In Section~\ref{conditional}, we discuss
forms of conditional interchangeability.  In Section~\ref{otherforms},
we discuss other forms of interchangeability that have appeared in the
literature.  In Section~\ref{classification}, we relate the various
forms of interchangeability. Finally, in Section~\ref{conclusion}, we
list topics that we plan to cover more fully in the expanded version
of this paper.

\section{Basic Interchangeability Concepts}
\label{concepts}

In this section, we review the various forms of interchangeability
originally introduced in \cite{freuder1991eliminating}.  We also
include a few new interchangeability concepts that directly relate to
the original ones.  Full interchangeability, the most basic form of
interchangeability, is defined as follows.

\vspace{0.3\baselineskip}
\noindent{\bf Full interchangeability (FI)} \cite{freuder1991eliminating} 
A value $a$ for variable $v$ is fully interchangeable with value $b$
iff every solution in which $v=a$ remains a solution when
$b$ is substituted for $a$ and vice-versa.  

\begin{figure}[!ht]
\begin{minipage}[t]{.30\textwidth}
\centerline{\includegraphics[scale=0.25]{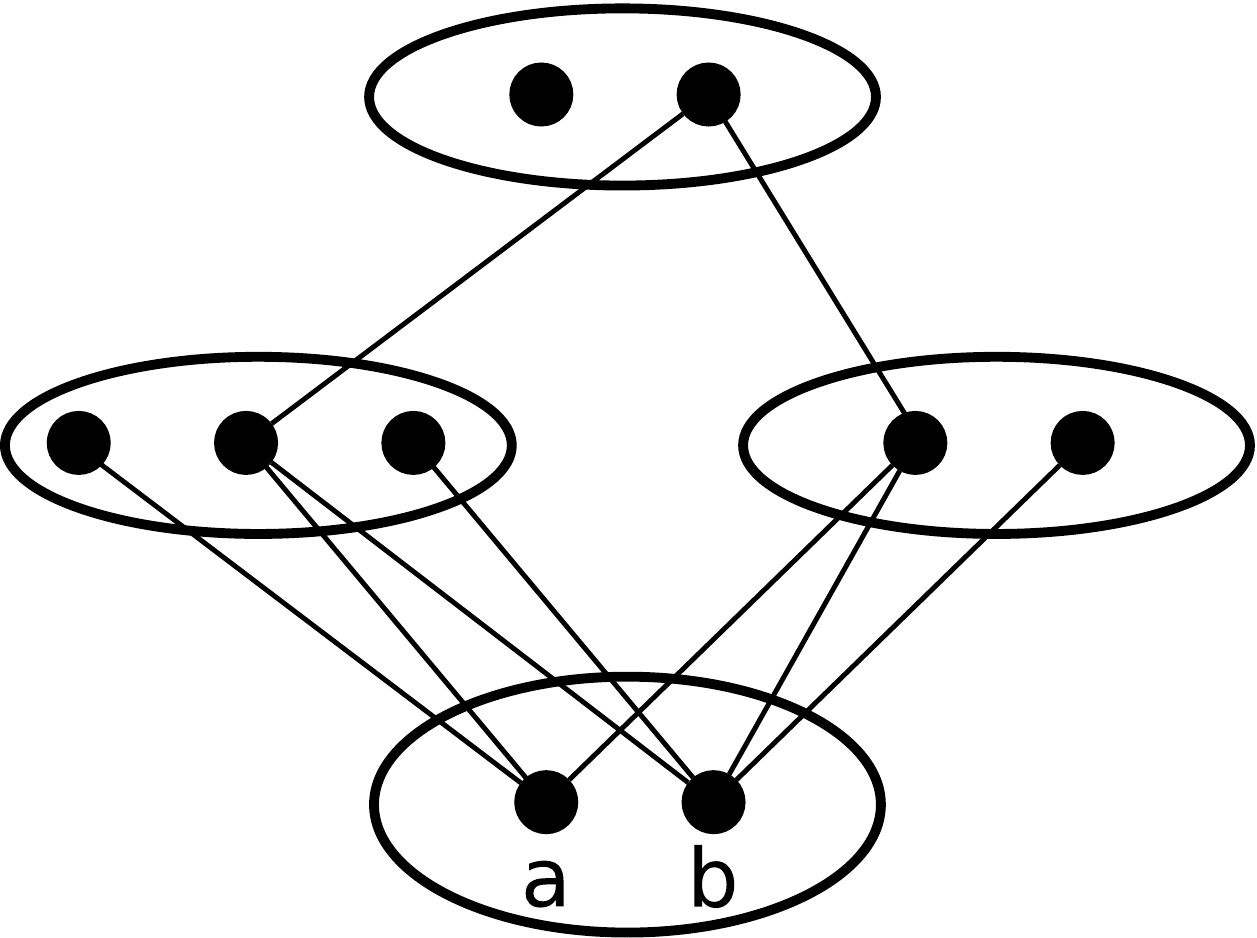}} \caption{\small FI: $a$
and $b$ are FI but not 3-I or NSub.}  \label{fig:FI} \end{minipage}
\hfil \begin{minipage}[t]{.30\textwidth}
\centerline{\includegraphics[scale=0.25]{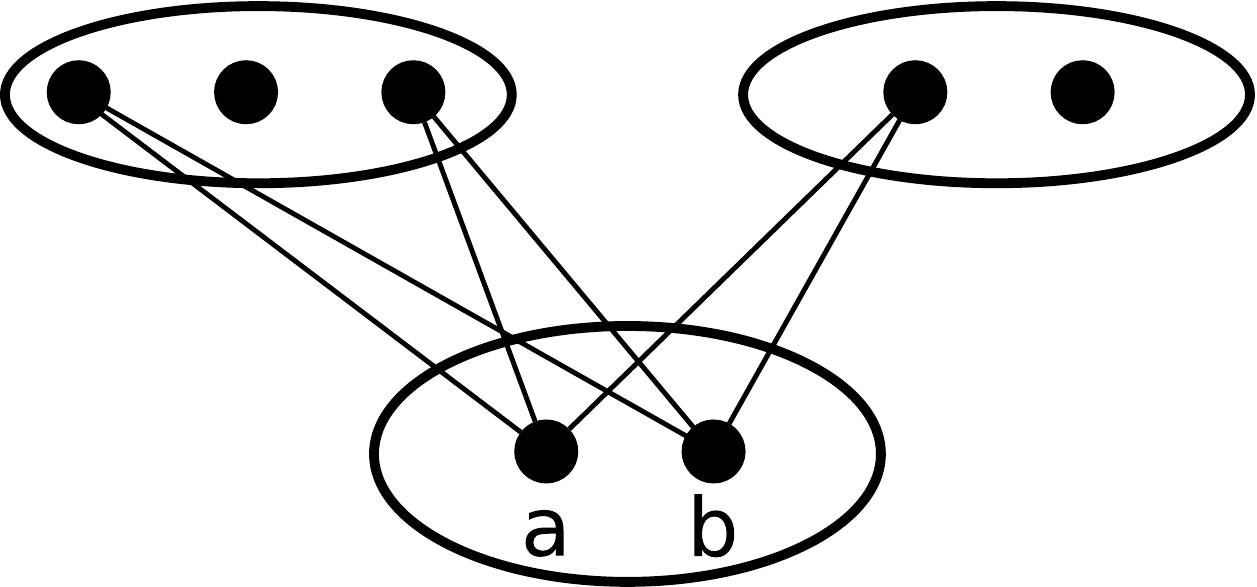}} \caption{\small NI: $a$
and $b$ are NI.}  \label{fig:NI} \end{minipage} \hfil
\begin{minipage}[t]{.30\textwidth}
\centerline{\includegraphics[scale=0.25]{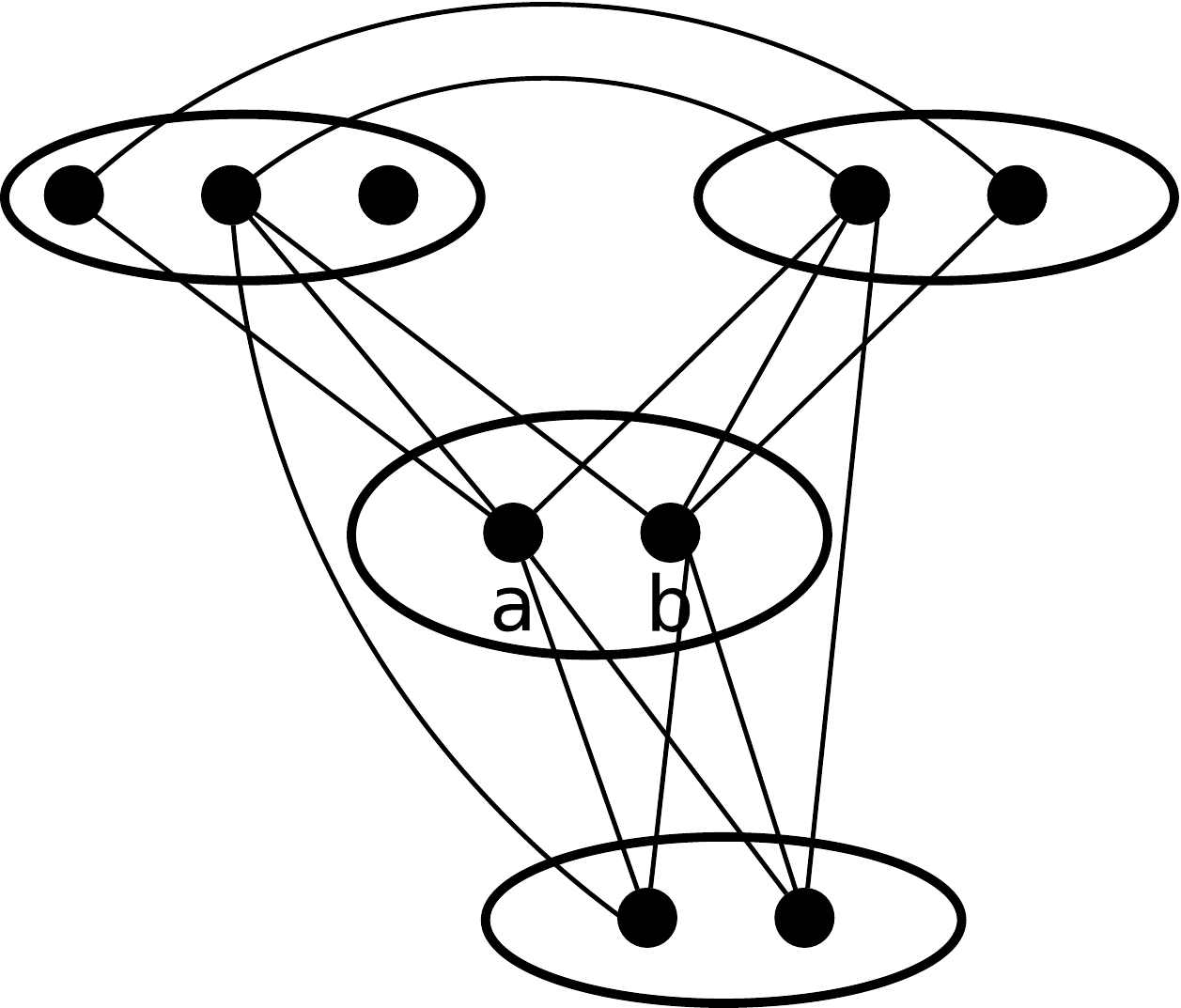}} \caption{\small KI: $a$
and $b$ are 3-I but not NI.}  \label{fig:KI} \end{minipage}
\end{figure}

If two values are interchangeable then one of them can be removed from
the domain, reducing the size of the problem; alternatively they can
be bundled together in a Cartesian product representation of
solutions.  Figures~\ref{fig:FI}, \ref{fig:NI} and ~\ref{fig:KI} show
examples of two values $a$ and $b$ that are FI (see below for
definition of 3-I and NSub). In our figures, a small solid circle
denotes a value in the domain of the variable represented by the
outline circle, and edges link consistent tuples.

Notice that FI is defined `at the solution level', which means that in
order to find all FI values for a given variable, one must account for
{\em all\/} constraints and may have to compute all solutions. Thus,
FI is a {\em global\/} property.  In \shortcite{benhamou1994study},
Benhamou defines the equivalent notion of `value symmetry in all
solutions' as {\em semantic symmetry}.  Hence, the terms `semantic'
and `global' are equivalent.  Because the detection of global forms of
interchangeability is likely to be intractable, Freuder introduced
{\em local\/} variants, which account only for the constraints defined
on a variable, that is, the neighborhood of the variable.  In
\shortcite{benhamou1994study}, Benhamou calls such relations  
{\em syntactic\/} symmetries.  Section~\ref{sec:local} discusses local
interchangeability.  Further, interchangeability is an equivalence
relation on the domain of the variable: interchangeable values are
equivalent.  Such equivalences may be rare in practice.  To remedy
this situation, Freuder proposed various extensions to the basic
concept, which are discussed in Sections~\ref{sec:weak}
and~\ref{sec:extended}.

In summary, one may think of interchangeability as a core concept
characterized as a relation between two values either at the solution
level (i.e., global or semantic) or in the neighborhood of the
variable (i.e., local or syntactic).  Also, the concepts may require
that interchangeable values be equivalent (i.e., strong), or not
`perfectly' so (i.e., weak or approximate).

{\em When comparing two forms of interchangeability $X$ and $Y$, we
say that $X \rightarrow Y$ iff any two values $a$ and $b$ that are
related by $X$ are also related by $Y$ but the converse does not
necessarily hold\footnote{$a$ and $b$ are two values or two partial
assignments over the same variables.},\/} regardless of whether $Y$ is
derived from $X$ by relaxing the conditions of $X$ (i.e., $Y$ is
weaker than $X$) or by `moving' from the local level to the global
level (i.e., syntactic to semantic). Note that when $X \rightarrow Y$,
$Y$ leads to greater problem reduction than $X$.

\subsection{Local forms of interchangeability}
\label{sec:local}

In general, the identification of a local interchangeability is
tractable because it focuses on the neighborhood of the
variable. Also, a given local form interchangeability usually implies
the corresponding global one.

\vspace{0.3\baselineskip}
\noindent{\bf Neighborhood interchangeability (NI)}
\cite{freuder1991eliminating} A value $a$ for variable $v$ is
neighborhood interchangeable with value $b$ iff for every constraint
on $v$, the values compatible with $v=a$ are exactly those compatible
with $v=b$. Values $a$ and $b$ are NI in Figure~\ref{fig:NI} but not
in Figures~\ref{fig:FI} or~\ref{fig:KI}.

Neighborhood interchangeable values for a given variable can be
detected by comparing the values in the variable's domain for
consistency to all variable-value pairs in the variable's neighborhood
and drawing a discrimination tree \cite{freuder1991eliminating}. At
the end of the process, the leaves of the discrimination tree are
annotated with the equivalence NI values for the variable. The
complexity of this process is $\mathcal{O}(n^2 d^2)$, where $n$ is the
number of variables and $d$ is the maximum domain size.
Alternatively, one can build a refutation tree, which proceeds by
splitting the domain of the variable
\cite{likitvivatanavong2008refutation}.  The lower bound of the
worst-case complexity of the refutation tree is smaller than that of
the discrimination tree.  However, it is not clear whether the
difference is meaningful in practice. Further, the discrimination tree
can be directly used to implement forward checking at no additional
cost \cite{BecEtc01}, but it is not clear yet whether or not the same
can be done with the refutation tree.

For non-binary constraints, neighborhood interchangeable values can be
detected by constructing non-binary discrimination trees for each
variable \cite{laletc05}. As described in \cite{laletc05}, the process
also allows the use of forward checking during search.  The complexity
to build a non-binary discrimination tree for a single variable is
$\mathcal{O}(n \,\mbox{\it deg}\, a^{k+1} (1-t))$, where $n$ is the
number of variables, {\it deg\/} is the maximum degree of a variable,
$a$ is the maximum domain size, and $t$ is the tightness of the
constraints, defined as the ratio of the number of forbidden tuples
over the number of all possible tuples.

\vspace{0.3\baselineskip}
\noindent{\bf K-interchangeability (KI)} 
\cite{freuder1991eliminating}
For $k\geq 2$, two values, $a$ and $b$ for a CSP variable $X$, are
\emph{k-interchangeable} iff $a$ and $b$ are fully interchangeable in
any subproblem of the CSP induced by $X$ and $(k-1)$ other variables.
Values $a$ and $b$ are 3-I in Figures~\ref{fig:NI} and~\ref{fig:KI}
but not in Figure~\ref{fig:FI}.

K-interchangeable values can be identified by a modification of the
discrimination-tree algorithm for NI.  The complexity of the process
is $\mathcal{O}(n^k d^k)$ \cite{freuder1991eliminating}.

\begin{theorem} {\em NI $\rightarrow$ KI $\rightarrow$ FI, see
  \cite{freuder1991eliminating}.}
\label{thm:NI-KI-PI}
\end{theorem}
For $2<i<j<|V|$, $i$-interchangeability is a sufficient but not
necessary condition for $j$-interchangeability. NI is
2-interchangeability and FI is $|V|$-interchangeability. Hence, NI
$\rightarrow$ KI $\rightarrow$ FI.  $a$ and $b$ in Figure~\ref{fig:FI}
are FI but not 3-I, and in Figure~\ref{fig:KI} they are 3-I but not
NI.

\subsection{Extended interchangeability: Weak forms}
\label{sec:weak}
Below, we discuss three weak forms of interchangeability introduced in
\cite{freuder1991eliminating} (i.e., subproblem interchangeability, partial
interchangeability, and substitutability) and a number of other
related concepts.  

\vspace{0.3\baselineskip}
\noindent{\bf Subproblem interchangeability (SPrI)} 
\cite{freuder1991eliminating}
Two values are \emph{subproblem interchangeable}, with respect to a
subset of variables $S$, iff they are fully interchangeable with
regards to the solutions of the subproblem of the CSP induced by $S$.

\begin{figure}[!ht]
  \begin{minipage}[t]{.32\textwidth}
  \centerline{\includegraphics[scale=0.25]{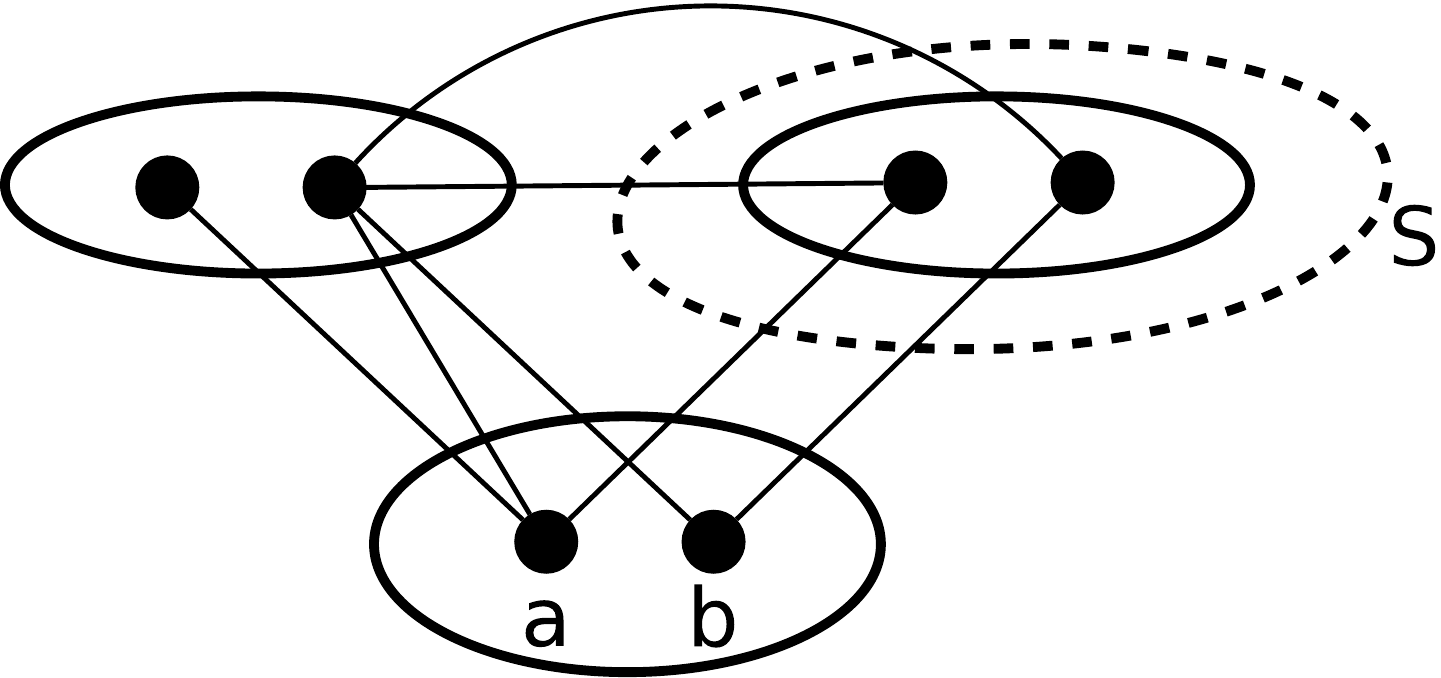}}
  \caption{\small PI: $a$ and $b$ are PI {\em wrt\/} $S$ but not Sub,
    FI, CtxDepI, NTI, or NPI {\em wrt\/} any subset.}
  \label{fig:PI} 
  \end{minipage}
  \hfil
  \begin{minipage}[t]{.30\textwidth}
  \centerline{\includegraphics[scale=0.25]{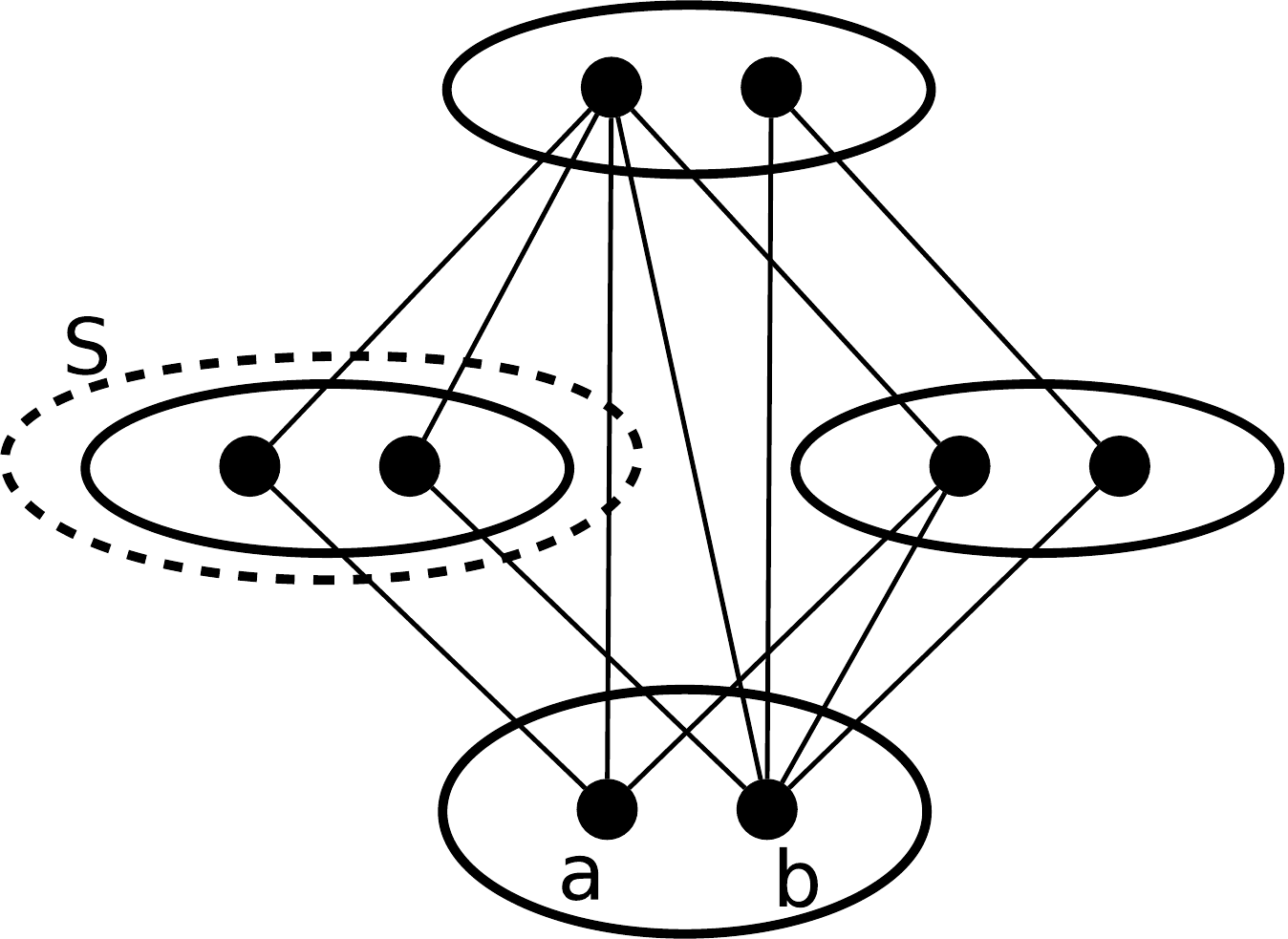}}
  \caption{\small PI: $a$ and $b$ are PI {\em wrt\/} $S$ but not Sub
    or SPrI {\em wrt\/} any subset of variables.}
  \label{fig:PI2} 
  \end{minipage}
  \hfil
  \begin{minipage}[t]{.30\textwidth}
  \centerline{\includegraphics[scale=0.25]{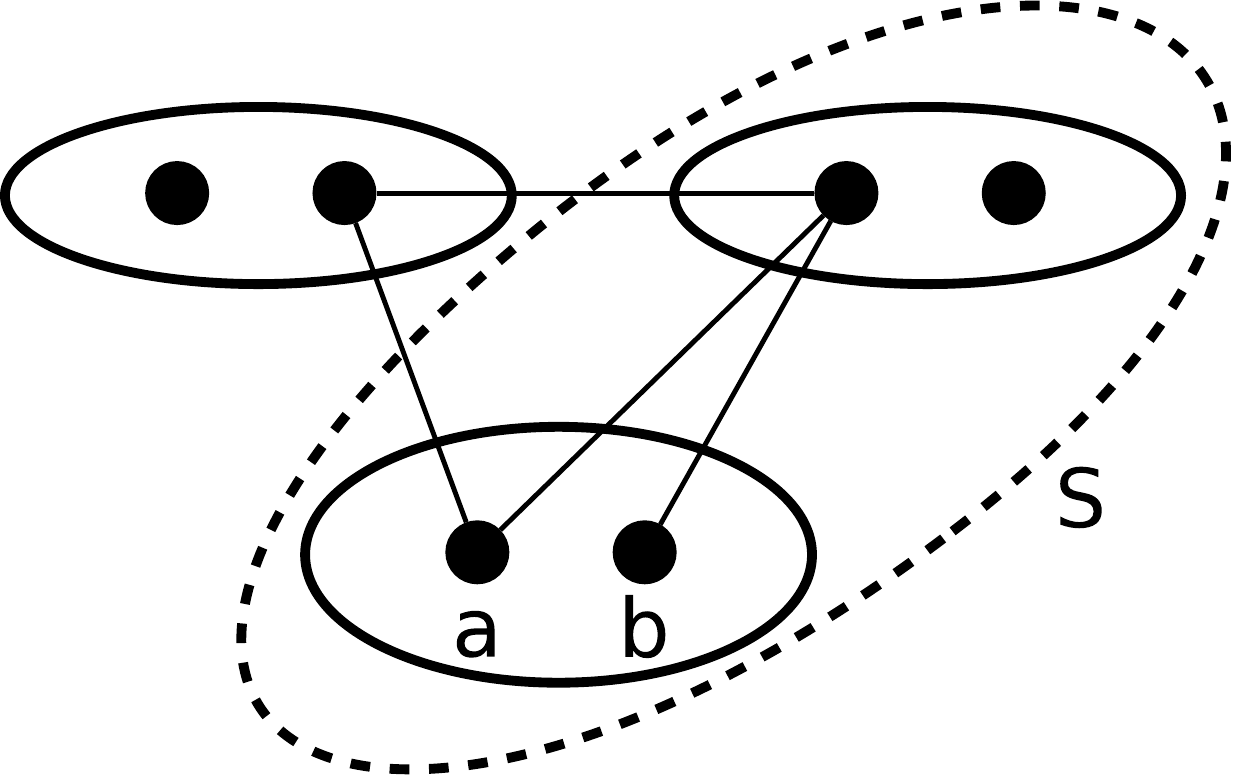}}
  \caption{\small SPrI: $a$ and $b$ are SPrI {\em wrt\/} $S$ but not
    PI {\em wrt\/} any subset of variables.}
  \label{fig:SI2} 
  \end{minipage}
  \end{figure}

\vspace{0.3\baselineskip}
\noindent{\bf Partial interchangeability (PI)} 
\cite{freuder1991eliminating} Two values are {\em partially interchangeable\/}
with respect to a subset $S$ of variables, iff any solution involving
one implies a solution involving the other with possibly different
values for variables in $S$.  In Figure~\ref{fig:PI}, $a$ and $b$ are
PI {\em wrt\/} $S$, shown with the dotted line.

\begin{theorem} {\em FI $\rightarrow$ PI.}
\end{theorem}
If $a$ and $b$ are FI, they are by definition PI with respect to any
subset of $V$.  In Figure~\ref{fig:PI}, $a$ and $b$ are PI {\em wrt\/}
to the subset $S$ but not FI.

\begin{theorem} {\em SPrI and PI are not comparable}\footnote{This theorem 
corrects Theorem~5 of \cite{freuder1991eliminating}.}.
\end{theorem}
In Figure~\ref{fig:PI2}, $a$ and $b$ are PI but not SPrI. In
Figure~\ref{fig:SI2}, $a$ and $b$ are SPrI but not PI.

\begin{figure}[!ht]
  \begin{minipage}[t]{.36\textwidth}
  \centerline{\includegraphics[scale=0.25]{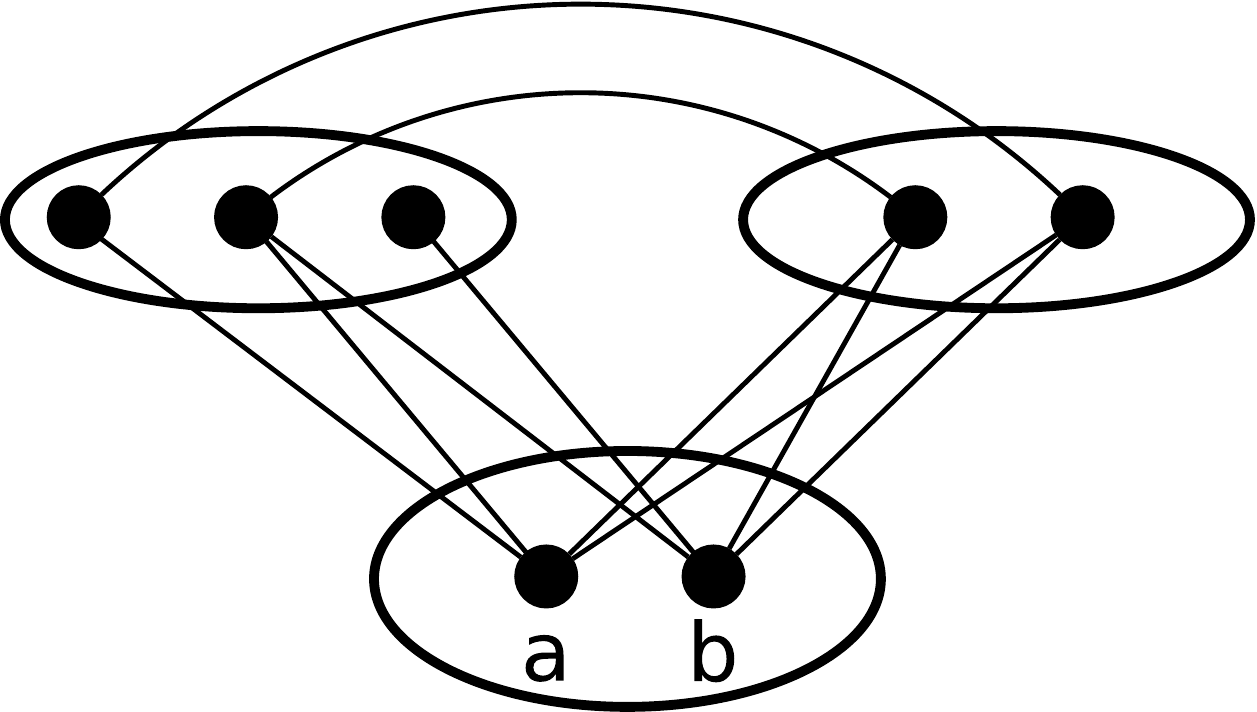}}
  \caption{\small Sub: $a$ is Sub, but not NSub, for $b$; $a$ and $b$ are not FI.}
  \label{fig:SUB} 
  \end{minipage}
  \hfil
  \begin{minipage}[t]{.30\textwidth}
  \centerline{\includegraphics[scale=0.25]{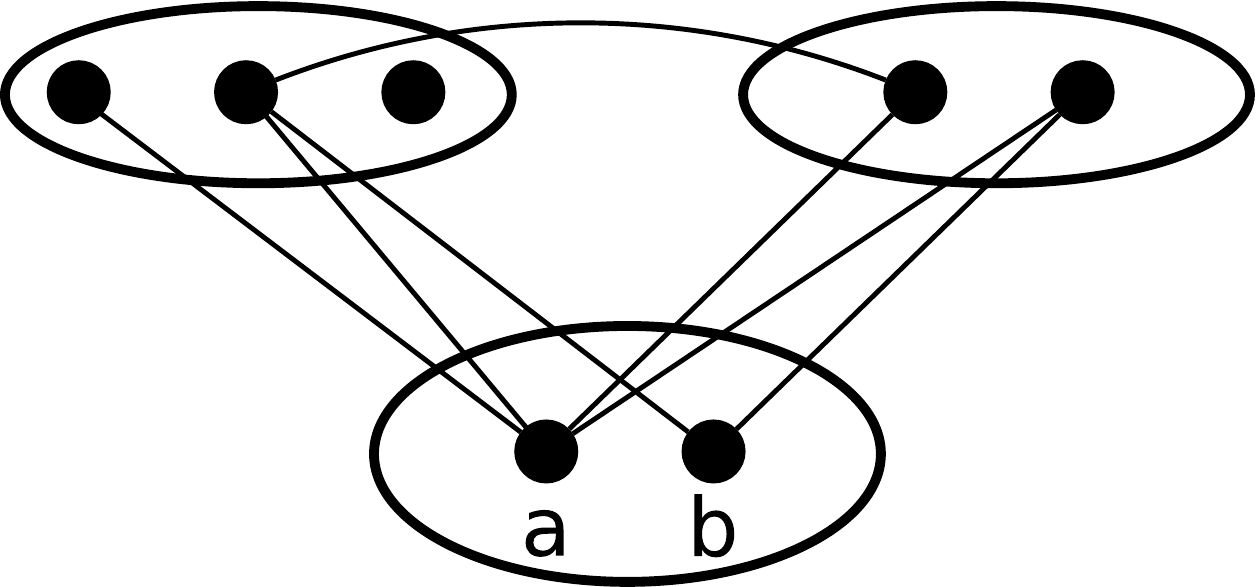}}
  \caption{\small $a$ is NSub for $b$ but $a$ and $b$ are not NI or FI.}
  \label{fig:NSub} 
  \end{minipage}
  \hfil
  \begin{minipage}[t]{.26\textwidth}
  \centerline{\includegraphics[viewport=0in 5.6in 3in 7.3in,clip=true,scale=0.4]{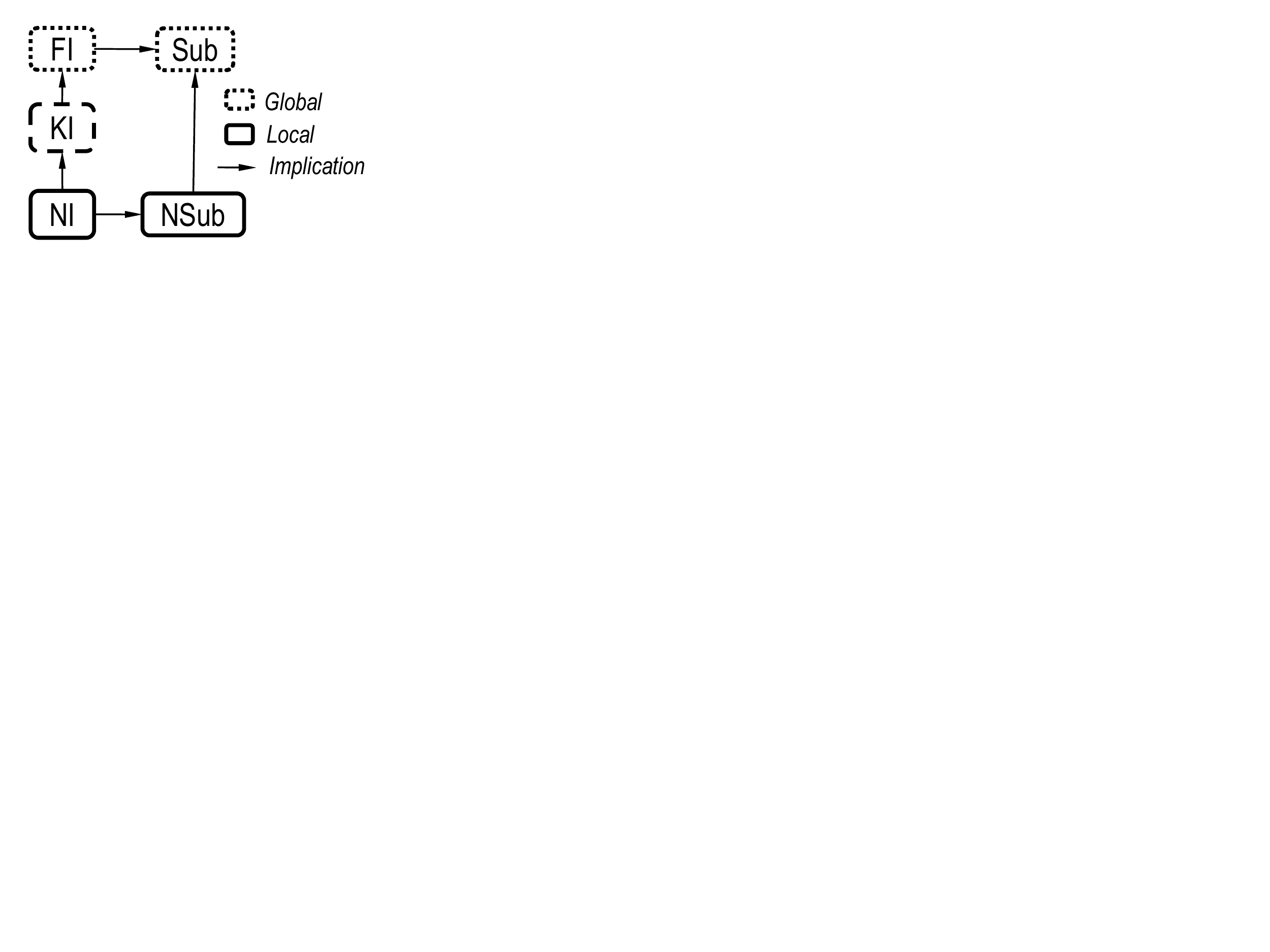}}
  \caption{\small Illustrating Theorems~\ref{thm:NI-KI-PI}, \ref{thm:FI-Sub}, and~\ref{thm:NI-NSub-Sub-FI}.}
  \label{fig:small-hasse} 
  \end{minipage}
  \end{figure}

\vspace{0.3\baselineskip}
\noindent{\bf Substitutability (Sub)} \cite{freuder1991eliminating}
 For two values $a$ and $b$ for variable $v$, $a$ is substitutable for
 $b$ iff every solution in which $v=b$ remains a solution when $b$ is
 replaced by $a$ but not necessarily vice-versa. Figure~\ref{fig:SUB}
 shows an example.

 Note that the concept of substitutability is related to that of
 dominance~\cite{bellicha1994csp}, which is used in the literature on
 symmetry breaking.

\begin{theorem} {\em FI $\rightarrow$ Sub.}
\label{thm:FI-Sub}
\end{theorem}
If $a$ and $b$ are FI, they are by definition mutually
substitutable. In Figure~\ref{fig:SUB}, $a$ is substitutable for $b$,
but $a$ and $b$ are not FI.

Again, because substitutable values are expensive to compute,
neighborhood substitutability (NSub) (with the obvious definition) is
computationally advantageous. In Figure~\ref{fig:NSub}, $a$ is NSub
for $b$.  In \shortcite{bellicha1994csp}, Bellicha et al.\ propose
{\sc NS-Closure}, an algorithm to enforce NSub.  It removes all of the
neighborhood substitutable values from the network.  It operates by
examining every pair of values $(a,b)$ in a variable's domain, trying
to find a {\it splitter\/} for the pair. A splitter for $(a,b)$ is a
value in the neighborhood of the variable that supports $a$ but not
$b$. If $(a,b)$ does not have a splitter, then $a$ can be removed from
the domain. The time complexity of the algorithm is $\mathcal{O}(m
d^3)$, where $m$ is the number of constraints and $d$ is the maximum
domain size.  The space complexity of storing the splitters is
$\mathcal{O}(n d^2)$, where $n$ is the number of variables.

\begin{theorem} {\em NI $\rightarrow$ NSub $\rightarrow$ Sub; FI
and NSub are not comparable.}
\label{thm:NI-NSub-Sub-FI}
\end{theorem}
  Figure~\ref{fig:small-hasse} illustrates this situation.  First
  consider NI $\rightarrow$ NSub.  Any NI values are mutually NSub by
  definition. In Figure~\ref{fig:NSub}, $a$ is NSub for $b$ but $a$
  and $b$ are not NI.  Now, consider NSub $\rightarrow$ Sub.  Given
  two values $a$ and $b$ for a variable, $a$ is NSub for $b$, the set
  of variable-value pairs supporting $b$ is a subset of the one
  supporting $a$. By moving to global substitutability, the sets of
  supports will only lose elements, however, the set of support of $b$
  will remain a subset of that of $a$. Figure~\ref{fig:SUB} shows an
  example where $a$ is Sub, but not NSub, for $b$.  In
  Figure~\ref{fig:FI}, $a$ and $b$ are FI but not NSub. In
  Figure~\ref{fig:NSub}, $a$ is NSub for $b$ but $a$, and $b$ are not
  FI.

\vspace{0.3\baselineskip}
\noindent{\bf Neighborhood Partial Interchangeability (NPI)} 
\cite{choueiry1998computation} Two values $b$ and $c$ for a variable
$v$ are NPI given a boundary of change $S$ (which includes $v$) iff,
for every constraint $C$ defined on the variables ($v$,$w$) where
$v\in S$, $w\notin S$, we have: $\{j|(b,j)$ satisfies $C\}=\{j|(c,j)$
satisfies $C\}$.

The NPI sets of a variable's domain can be detected by modifying the
discrimination tree algorithm of NI to a joint discrimination tree
(JDT) by considering the neighborhood of a set of variables instead of
the neighborhood of a single variable as done in the discrimination
tree \cite{choueiry1998computation}.  The complexity of the algorithm
to build a JDT for a single variable is $\mathcal{O}(s(n-s)d^2)$ and
the space complexity for the tree is $\mathcal{O}((n-s)d)$, where $n$
is the number of variables, $s$ is the size of the given set, and $d$
is the size of the domain.

\begin{figure}[!ht]
  \begin{minipage}[t]{.45\textwidth}
    \centerline{\includegraphics[scale=0.25]{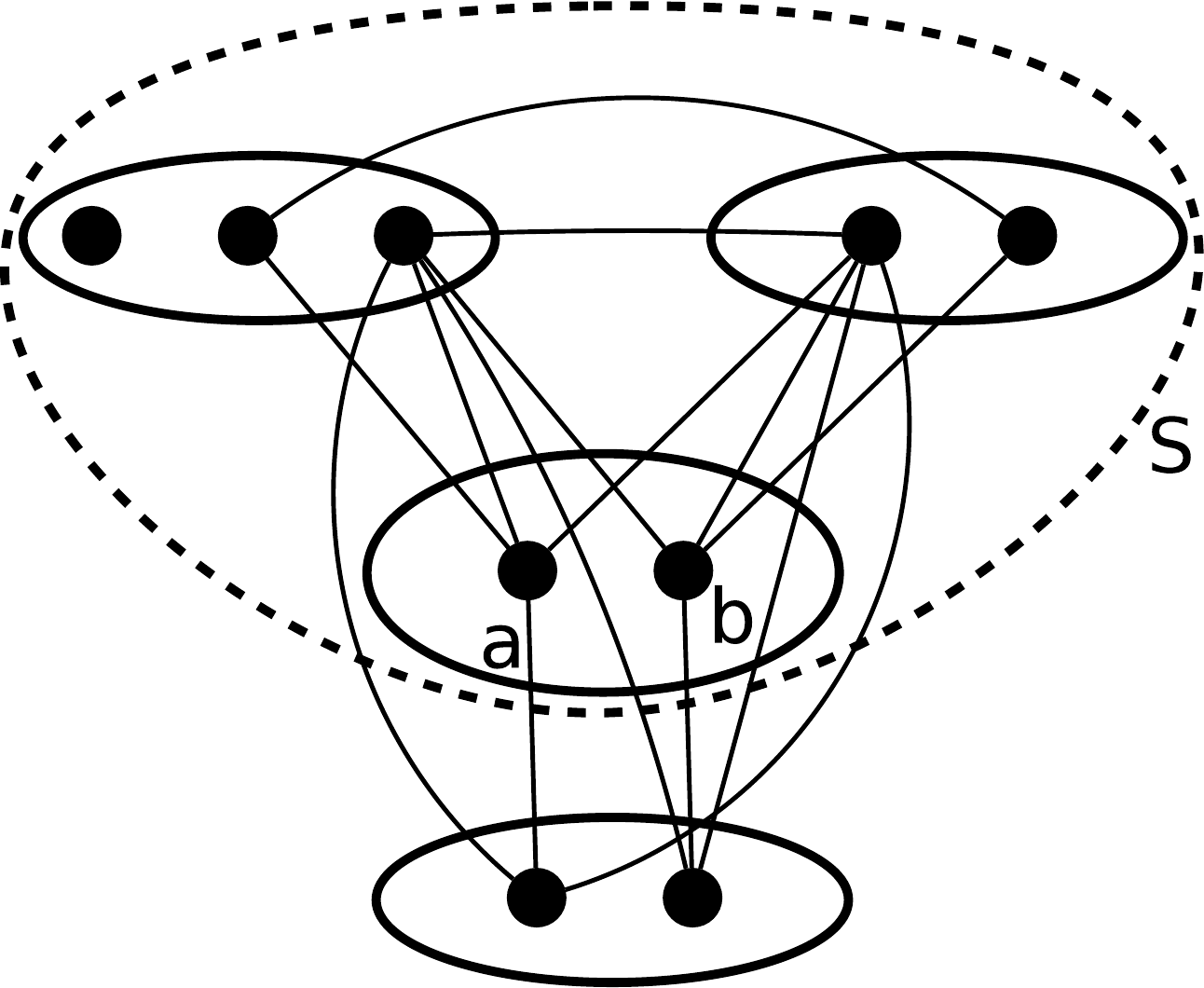}}
    \caption{\small SPrI: $a$ and $b$ are SPrI {\em wrt\/} $S$ but not
      NPI {\em wrt\/} to any subset or Sub.}
    \label{SI}
  \end{minipage}
  \hfil
  \begin{minipage}[t]{.50\textwidth}
    \centerline{\includegraphics[scale=0.25]{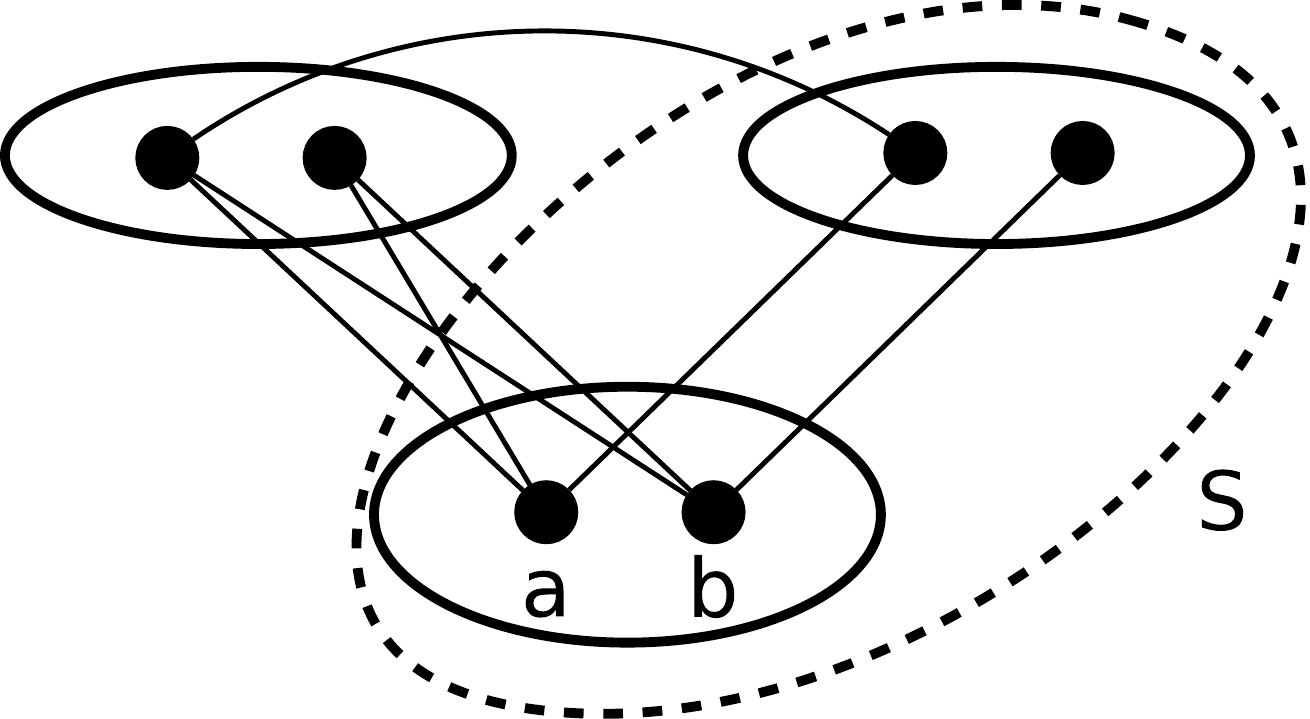}}
    \caption{\small NPI: $a$ and $b$ are NPI {\em wrt\/} $S$ but not PI
      {\em wrt\/} to any subset, SUB, FI or NTI.}
    \label{fig:NPI}
  \end{minipage}
\end{figure}

\begin{theorem} {\em NPI and PI are not comparable.}\footnote{This
    theorem corrects \cite{choueiry1998computation}, which states that
    NPI implies PI. This error was mentioned in \cite{neagu2005approximating}.}
\end{theorem}
In Figure~\ref{fig:PI}, $a$ and $b$ are PI but not NPI. In
Figure~\ref{fig:NPI}, they are NPI but not PI.

\begin{theorem} {\em NPI $\rightarrow$ SPrI.}
\end{theorem}
If $a$ and $b$ are NPI outside the boundary of change $S$, then they
are NI in the subproblem induced by $V\setminus S$. If they are NI in
the subproblem, then they are also FI, and therefore SPrI in the
subproblem induced by $V\setminus S$. Figure~\ref{SI} shows an example
where the converse does not hold.

\vspace{0.3\baselineskip}
\noindent{\bf Directional Interchangeability (DirI)} 
\cite{naanaa-directional}
Two values $a$ and $b$ in the domain of a variable $X$ are DirI with
respect to a variable ordering of the variables iff they have the same
preceding support set: $\{c\ |\ (a,c)\in C_{XY}$ and $Y\prec X\}=\{c\
|\ (b,c)\in C_{XY}$ and $Y\prec X\}$ .

\begin{theorem} {\em NPI $\equiv$ DirI.}
\end{theorem}
 If $a$ and $b$ for a variable $X$ are NPI {\em wrt\/} a boundary of
 change $S$, then they are DirI {\em wrt\/} any variable ordering such
 that $\forall Y\in S$, $X\prec Y$.

\ig{Directional interchangeability values can be detected during search
using the algorithm for detecting NI values, which generates
discrimination trees.  In order to avoid the overhead of generating
discrimination trees during search, So, they introduce a new algorithm
which has the same time complexity for detecting DynI for a single
variable as detecting NI for a single variable, $\mathcal{O}(n d^2)$,
where $n$ is the number of variables and $d$ is the maximum domain
size, but does not have the overhead of creating a discrimination tree
at each step.  This algorithm for finding the DynI values can then be
combined with forward checking or maintaining arc consistency
algorithms during search \cite{naanaa-directional}.}

\vspace{0.3\baselineskip}
\noindent{\bf Directional Substitutability (DirSub)} 
\cite{naanaa2007substitutability,naanaa2009domain} Value $a$ is DirSub for value $b$  for a
variable $X$ with respect to a variable ordering of the variables iff
the preceding support set of $b$ is a subset of that of $b$: $\{c\ |\
(b,c)\in C_{XY}$ and $Y\prec X\}\subseteq\{c\ |\ (a,c)\in C_{XY}$ and
$Y\prec X\}$.

\begin{theorem} {\em DirI $\rightarrow$ DirSub.} 
\end{theorem}
If $a$ and $b$ of variable $X$ are DirI {\em wrt\/} a given variable
ordering, then, by definition, the preceding support sets for $a$ and
$b$ are the same and consequently subsets of one another.

\vspace{0.3\baselineskip}
\noindent{\bf Neighborhood Interchangeability Relative to a Constraint (NI$_C$)} 
\cite{haselbock1993exploiting} Two values are NI$_C$ relative to a constraint 
$C$ iff they are NI in the problem induced by the variables in the
scope of $C$.

NI$_C$ values for variable $v$ can be detected by restricting the
discrimination tree to the considered constraint.  As a result, the
time complexity of finding all NI$_C$ sets is $\mathcal{O}(e k a^k)$,
where $e$ is the number of constraints, $k$ is the maximum arity, and
$a$ is the maximum domain
size. In~\shortcite{haselbock1993exploiting}, Haselb\"{o}ck modified
the usual {\sc Revise\/} procedure for lookahead to exploit the
(statically computed) NI$_C$ sets during search, yielding a solution
bundle. The time complexity of the new {\sc Revise\/} procedure is
thus reduced to $\mathcal{O}(a'^2)$, where $1 \leq a' \leq a$.  In
\cite{BecEtc01}, it was shown that the resulting bundles are never
`thinner' than those obtained in \cite{BenFre92}, and never `fatter'
than those obtained by those obtained in~\cite{HubbeF92}, which in
turn are equivalent to those obtained by~\cite{BecEtc01}.

\vspace{0.3\baselineskip}
\noindent{\bf Neighborhood Substitutability Relative to a Constraint (NSub$_C$)} 
\cite{boussemart2004support}
Two values are NSub$_C$ relative to a constraint $C$ iff they are NSub
in the problem induced by the variables in the scope of $C$.

\begin{theorem} {\em NPI $\rightarrow$ NI$_C$ $\rightarrow$ NS$_C$.}
\end{theorem}
First consider NPI $\rightarrow$ NI$_C$. If for variable $X$, $a$ and
$b$ are NPI, then for every constraint $C$ between $X$ and a variable
outside of the boundary of change, $a$ and $b$ are NI$_C$.  NI$_C$
$\rightarrow$ NS$_C$ follows directly from the definition.

\subsection{Other extended forms of interchangeability}
\label{sec:extended}
Other extended forms that were initially proposed are:
meta-interchangeability, dynamic interchangeability, and functional
interchangeability.  

\vspace{0.3\baselineskip}
\noindent{\bf Meta-interchangeability (MI)}
\cite{freuder1991eliminating} By grouping variables into
`meta-variables', or values into `meta-values', we can introduce
interchangeability into higher level `meta-problem' representations of
the original CSP.

Values may become interchangeable or substitutable during backtrack
search after some variables have been instantiated, so even a problem
with no interchangeable values may exhibit interchangeability under
some search strategy.

\vspace{0.3\baselineskip}
\noindent{\bf Dynamic Neighborhood Interchangeability
  (DynNI)}\footnote{Dynamic Interchangeability (DynI) property was
  incorrectly characterized as Dynamic Neighborhood Partial
  Interchangeability (DNPI) in
  \cite{beckwith2001dynamic,Choueiry:sara02,lal-constraint,laletc05}.}
\cite{beckwith2001dynamic} Two values $a$ and $b$ for variable $X$ are
DynNI with respect to a set $A$ of variable assignments iff they are
NI in the subproblem induced by $A\cup\{X\}$ .

\begin{theorem} NI $\rightarrow$ DynNI.
\end{theorem}
Consider values $a$ and $b$ for a variable $v$ that are NI, assume $a$
and $b$ are not DynNI. Then, for an assignment for the subset of
variables $S$, either $a$ and $b$ are not NI in the problem induced by
$V\setminus S$, or one of $a$ or $b$ is deleted. The former case is
impossible because $a$ and $b$ have the same set of supports in the
original problem, and thus must have the same supports after the
assignments. The latter case is also impossible because $a$ and $b$
having the same support sets, if $a$ loses all its supports in a
neighboring variable, then $b$ also loses all supports because the
support sets are the same.

\vspace{0.3\baselineskip}
\noindent{\bf Full Dynamic Interchangeability (FDynI)} \cite{prestwich2004full}
A value $a$ for variable $v$ is dynamically interchangeable for $b$
with respect to a set $A$ of variable assignments iff they are fully
interchangeable in the subproblem induced by $A$.

\begin{theorem} {\em DynNI $\rightarrow$ FDynI.}
\end{theorem}
If $a$ and $b$ are DynNI, then $a$ and $b$ are consistent with the
same set of values in the assignment $A$.  They are also NI relative
to the variables in the problem induced by $A$ that are not yet
assigned and, consequently, are FI.

\vspace{0.3\baselineskip}
\noindent{\bf Functional interchangeability} 
\cite{freuder1991eliminating}
Let $S_{a|X}$ be the set of solutions including value $a$ for variable
$X$.  Two values $a$ for $X$ and $b$ for $Y$ are \emph{functionally
interchangeable} iff there exists functions $f$ and $f'$ such that
$f(S_{a|X})=S_{b|Y}$ and $f(S_{b|Y})=S_{a|X}$.

Two values $a$ and $b$ for a variable are {\em isomorphically
  interchangeable\/} \cite{freuder1991eliminating} iff there exists a
1-1 function $f$ such that $b=f(a)$ and for any solution $S$ involving
$a$, $\{f(v)\ |\ v\in S\}$ is a solution. Also for any solution $S$
involving $b$, $\{f^{-1}(v)\ |\ v\in S\}$ is a solution.

In the longer version of this paper, we compare functional and
isomorphic interchangeability with the definitions of symmetry
introduced in \cite{benhamou1994study,cohen2006symmetry}.

\section{Conditional Forms of Interchangeability} 
\label{conditional}

Conditions can be added to a CSP in the form of constraints that
further constrain the problem. In problems with little
interchangeability, such conditions can be imposed to increase the
interchangeability among the variable values.  In
\shortcite{zhang2004conditional}, Zhang and Freuder introduced and
studied conditional interchangeability, conditional substitutability,
conditional neighborhood interchangeability and conditional
neighborhood substitutability.

\vspace{0.3\baselineskip}
\noindent{\bf Conditional Interchangeability (ConI)} 
\cite{zhang2004conditional}
Two values $a$ and $b$ of variable $v$ are ConI under a condition imposed by a
set of additional constraints iff they are FI in the problem with the
additional constraints. 

Similarly Conditional Neighborhood Interchangeability (ConNI),
Conditional Substitutability (ConSub), and Conditional Neighborhood
Substitutability (ConNSub) are defined by \cite{zhang2004conditional}
where a problem is NI, Sub and NSub respectively given a set of
conditions.

\begin{theorem} {\em (ConNI $\rightarrow$ ConI $\rightarrow$ ConSub), 
(ConNI $\rightarrow$ ConNSub $\rightarrow$ ConSub), and ConI and
ConNSub are not comparable.}
\end{theorem}
For ConNI $\rightarrow$ ConI and ConNSub $\rightarrow$ ConSub, see
\cite{zhang2004conditional}.  Consider the local forms: 
ConNI $\rightarrow$ ConNSub.  For the same set of additional
constraints, if $a$ and $b$ are ConNI in the original problem, they
are NI in the problem with the additional constraints. Hence, they are
also NSub in the problem with the additional constraints, and ConNSub
in the original problem.  The proof for the global forms (i.e., ConI
$\rightarrow$ ConSub) is similar.  Similar to the non-comparability of
FI and NSub (see Theorem~\ref{thm:NI-NSub-Sub-FI}), ConI and ConNSub
can be shown to be not comparable.

\section{Other Forms of Interchangeability} \label{otherforms}

In this section we review other forms of interchangeability that have
appeared in the literature.

\begin{figure}[!ht]
  \begin{minipage}[t]{.30\textwidth}
  \centerline{\includegraphics[scale=0.25]{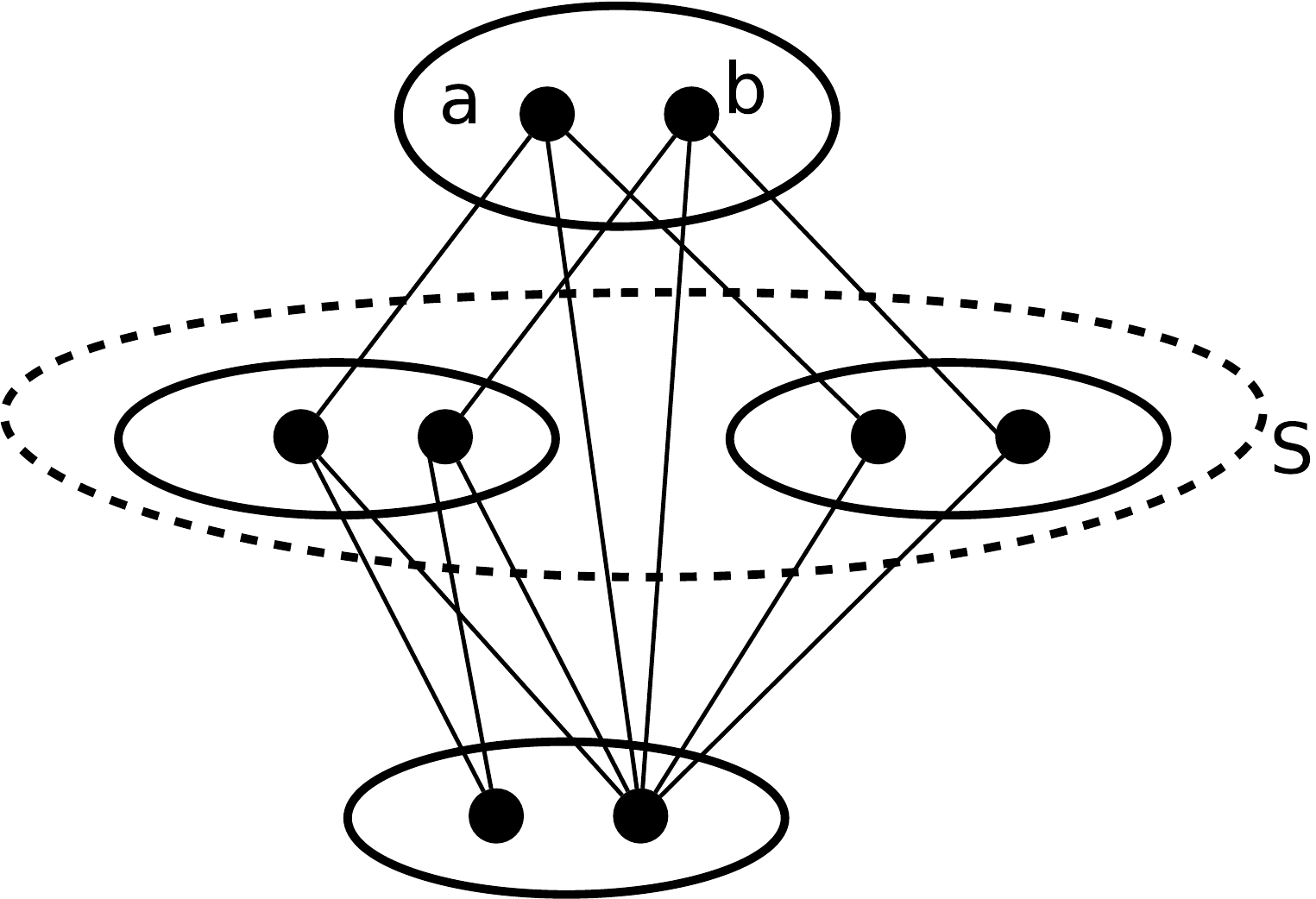}}
  \caption{\small NTI: $a$ and $b$ are NTI but not FDynNSub.}
  \label{fig:NTI} 
  \end{minipage}
  \hfil
  \begin{minipage}[t]{.36\textwidth}
  \centerline{\includegraphics[scale=0.25]{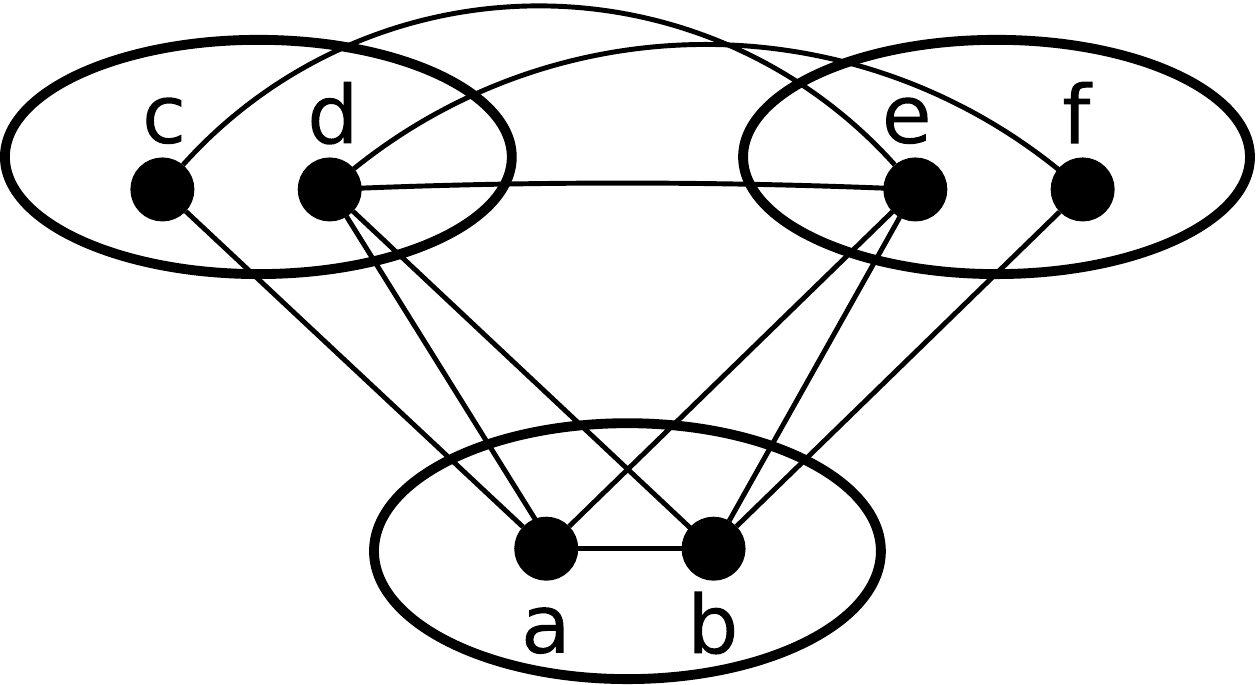}}
  \caption{\small CtxDepI: $a$ and $b$ are CtxDepI (clique $\{a,b,d,e\}$) but not Sub, FI, or PI.}
  \label{fig:CDI} 
  \end{minipage}
  \hfil
  \begin{minipage}[t]{.28\textwidth}
  \centerline{\includegraphics[scale=0.25]{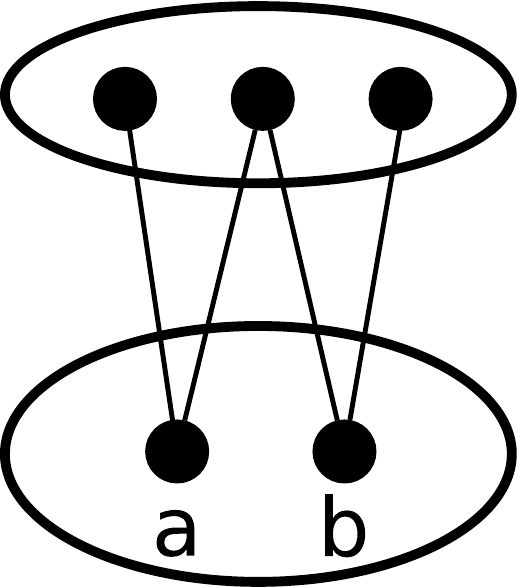}}
  \caption{\small GNSub: $a$ and $b$ are GNSub but not Sub.}
  \label{fig:GNS} 
  \end{minipage}
  \end{figure}

\vspace{0.3\baselineskip}
\noindent{\bf Neighborhood Tuple Interchangeability (NTI)} 
\cite{neagu1999constraint}. 
Values $a$ and $b$ for variable $v$ are NTI with respect to a set of
variables $S$ if for every consistent tuple $t$ of value assignments
to $S\cup\{v\}$ where $x=a$ there is another consistent tuple $t'$
where $v=b$ such that $t$ and $t'$ are consistent with the same value
combinations for variables outside of $S$. Additionally, the same
condition must hold when $a$ and $b$ are exchanged.
Figure~\ref{fig:NTI} shows an example.

The algorithm proposed in \cite{neagu2005approximating} to detect NTI
values determines the smallest set $S$ using discrimination trees.
The complexity of detecting NTI values is $\mathcal{O}((n^{s_{max}}
s_{max} (n - s_{max}) d^4)$, where $n$ is the number of variables, $d$
is the maximum domain size, and $s_{max}$ is the maximum size of all
possible dependent sets in the neighborhood of the variable.

\begin{theorem} {\em NI $\rightarrow$ NTI $\rightarrow$ PI and NTI $\rightarrow$ NPI.}
\end{theorem}
First, consider NI $\rightarrow$ NTI.  Given values $a$ and $b$ that
are NI for a variable, for every consistent tuple $t$ with $a$ there
is a tuple $t'$ that only differs from $t$ with $a$ replaced with
$b$. Hence $t$ and $t'$ are consistent with the same value
combinations. Figure~\ref{fig:NTI} gives an example where the converse
does not hold.  For (NTI $\rightarrow$ PI) and (NTI $\rightarrow$
NPI), see \cite{neagu2005approximating}.  Figure~\ref{fig:PI} gives an
example where PI $\not\rightarrow$ NTI, and Figure~\ref{fig:NPI} gives
an example where NPI $\not\rightarrow$ NTI.

In \shortcite{wilson2005decision}, Wilson described a new approach to
computation in a semiring-based system based on semiring-labeled
decision diagrams (SLDDs).  He defines forward neighborhood
interchangeability (ForwNI) and uses it for merging nodes in SLDDs,
hence compacting the search space.  During search, ForwNI takes into
account constraints that apply to instantiated and uninstantiated
variables.

\vspace{0.3\baselineskip}
\noindent{\bf Forward Neighborhood Interchangeability (ForwNI)} 
\cite{wilson2005decision}
Given a subset of variables $U\subset V$, two assignments $u$ and $u'$
to a set of variables $U$ are said to be ForwNI if for all constraints
$c\in C$ such that $scope(c)\cap (V\setminus U)\ne\emptyset$ and
$scope(c)\cap U\ne\emptyset$, $\Pi_{V\setminus U}\{t\in c \,|\,
\Pi_U(t)=u\}=\Pi_{V\setminus U}\{t\in c \,|\,
\Pi_U(t)=u'\}$.

\begin{theorem} {\em NTI $\rightarrow$ ForwNI}.
\end{theorem}
If $a$ and $b$ for variable $X$ are NTI with respect to set of
variables $S$, then for every assignment $t$ to $S\cup\{X\}$ where
$X=a$ there is another consistent tuple $t'$ where $X=b$ such that $t$
and $t'$ are consistent with the same value combinations for variables
outside of $S$. Hence, the set of tuples consistent with $t$ is the
same for $t'$ when projected on $V\setminus S$. Therefore, the
assignments $\Pi_S(t)$ and $\Pi_S(t')$ are ForwNI.

Tuple substitutability is a global form of ForwNI:

\vspace{0.3\baselineskip}
\noindent{\bf Tuple Substitutability (TupSub)} 
\cite{jeavons1994cp} Two assignments $A$ and $B$ to a set of
variables $R$ are TupSub iff $\Pi_{V\setminus
R}(\sigma_B(Sol))\subseteq\Pi_{V\setminus R}(\sigma_A(Sol))$, where
$Sol$ is the set of all solutions to the problem.
\begin{theorem} {\em ForwNI $\rightarrow$ TupSub.}
\end{theorem}
If two assignments $u$ and $u'$ are ForwNI, then for every solution in
which $u$ participates, $u$ can be substituted with $u'$ because they
have the same supports in every constraint that links the scope of $u$
to the rest of the problem. Therefore, the assignments $u$ and $u'$
are interchangeable and consequently substitutable.

\begin{theorem} {\em DynNI $\rightarrow$ ForwNI}.
\end{theorem}
If $a$ and $b$ for variable $X$ is DynNI {\em wrt\/} a set $A$ of
assignments, then any two assignments $u$ and $u'$ in $A\cup \{X\}$,
such that $\Pi_X(u)=a$ and $\Pi_X(u')=b$, have the same set of support
tuples because $a$ and $b$ are NI in $V\setminus A$. Therefore, $u$
and $u'$ are ForwNI.

\vspace{0.3\baselineskip}
\noindent{\bf Full Dynamic Substitutability (FDynSub)} 
\cite{prestwich2004fullsub}
A value $a$ for variable $v$ is dynamically substitutable with value
$b$ with respect to a set $A$ of variable assignments iff $a$ is fully
substitutable for $b$ in the subproblem induced by $A$.

\begin{theorem} {\em FDynI $\rightarrow$ FDynSub, follows directly from the
  definition.} 
\end{theorem}

\begin{theorem} {\em Sub $\rightarrow$ FDynSub.}
\end{theorem}
Consider value $a$ Sub for $b$ for variable $v$, the set of values
supporting $b$ is a subset of the set of values supporting $a$. Given
an assignment of variables in FDynSub, the sets of supports will only
lose elements, hence set of values supporting $b$ will remain a
subset of the set of values supporting $a$.

\begin{theorem} {\em FDynSub $\rightarrow$ ConNSub.}
\end{theorem}
If $a$ and $b$ are FDynSub, then a set of constraints can be
constructed for the original problem that removes all but the assigned
values in the variables that are assigned in FDynSub. In the new
problem resulting from adding those constraints, $a$ and $b$ are Sub.
Moreover, $a$ and $b$ are NSub because all the values in the
neighborhood that are not part of a solution are eliminated by the
added constraints.

\begin{theorem} {\em TupSub and  FDynSub are not comparable.}
\end{theorem}

\vspace{0.3\baselineskip}
\noindent{\bf Context dependent interchangeability (CtxDepI) } 
\cite{weigel1996context} 
Values $a$ and $b$ for a CSP variable $X$ are CtxDepI,
iff there exists a solution clique in the modified microstructure of
the CSP that contain both nodes ($X,a$) and ($X,b$). The modified
microstructure of the CSP is the original microstructure with edges
added between values of the same variable.
Figure~\ref{fig:CDI} shows an example.

\begin{theorem} {\em CtxDepI $\equiv$ FDynI.}
\end{theorem}
Connecting two CtxDepI values $a$ and $b$ for a CSP variable in the
micro-structure of the CSP yields a solution clique with $a$ and
$b$. By assigning the values in the clique to the variables, we obtain
an assignment set $A$ where $a$ and $b$ are fully interchangeable in
the subproblem induced by $A$. Conversely, if $a$ and $b$ are FDynI
with respect to an assignment set $A$, then there is a solution clique
in the modified micro-structure with $a$, $b$ and all the values in
$A$.

\begin{theorem} {\em FI $\rightarrow$ CtxDepI, FDynI.}
\end{theorem}
If $a$ and $b$ are FI for a variable $v$, then a solution with $v=a$
yields another solution when replacing $a$ with $b$. Thus, by
connecting $a$ and $b$ in the micro-structure, we obtain a solution
clique with $a$ and $b$. Figure~\ref{fig:CDI} shows an example where
the converse does not hold.

\vspace{0.3\baselineskip}
\noindent{\bf Generalized Neighborhood Substitutability (GNSub)} \cite{chmeiss2003neighborhood}
Two values of a variable are GNSub iff they share at least one support
with respect to each neighboring variable. Figure~\ref{fig:GNS} shows
an example.

\begin{figure}[!ht]
  \begin{minipage}[t]{.32\textwidth}
  \centerline{\includegraphics[scale=0.25]{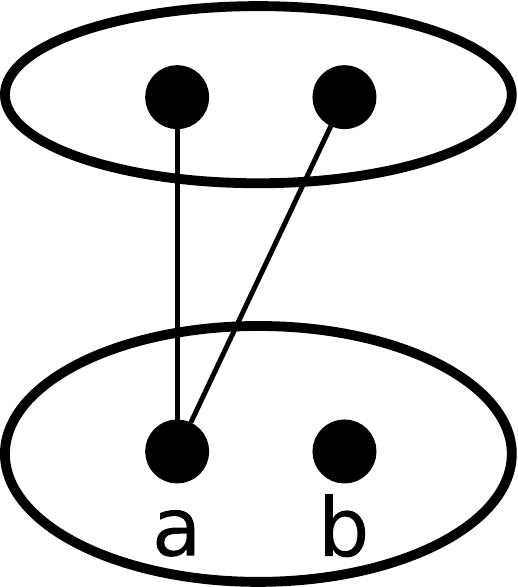}}
  \caption{\small NSub: $a$ and $b$ are NSub but not ConNI or GNSub.}
  \label{NSUB2}
  \end{minipage}
  \hfil
  \begin{minipage}[t]{.30\textwidth}
  \centerline{\includegraphics[scale=0.25]{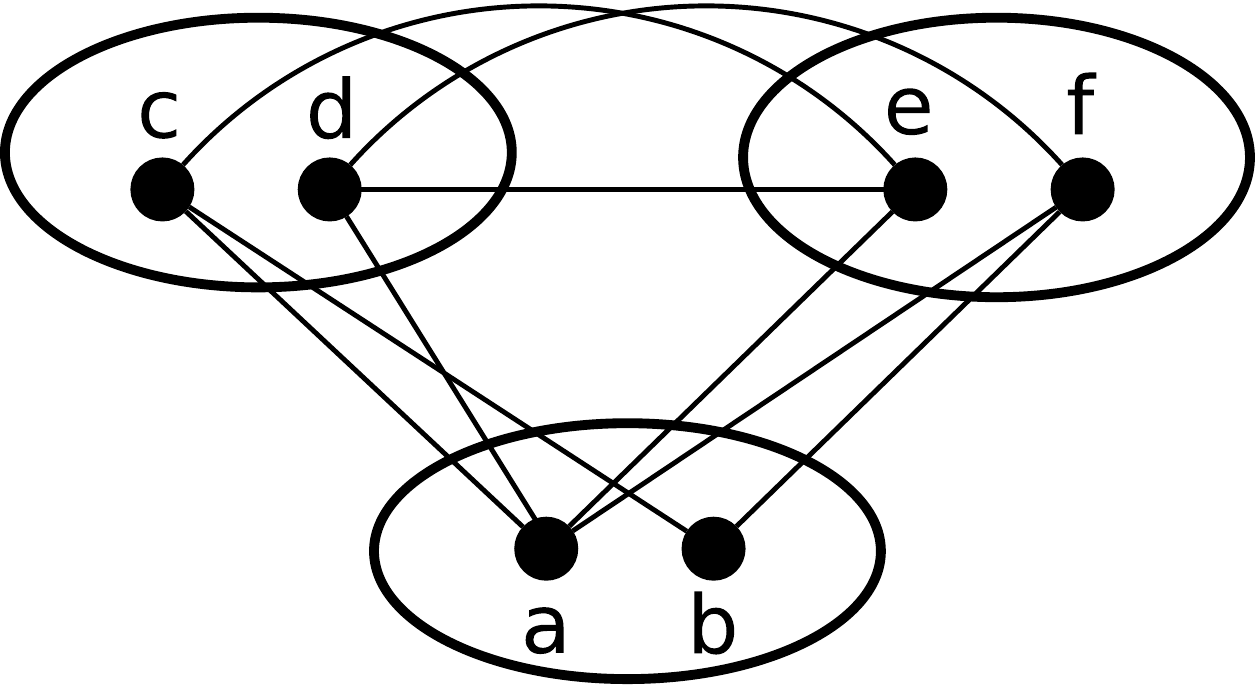}}
  \caption{\small GNSub: $a$ and $b$ are GNSub but not FDynI or
    CtxDepI.}
  \label{GNS2}
  \end{minipage}
  \hfil
  \begin{minipage}[t]{.30\textwidth}
  \centerline{\includegraphics[scale=0.25]{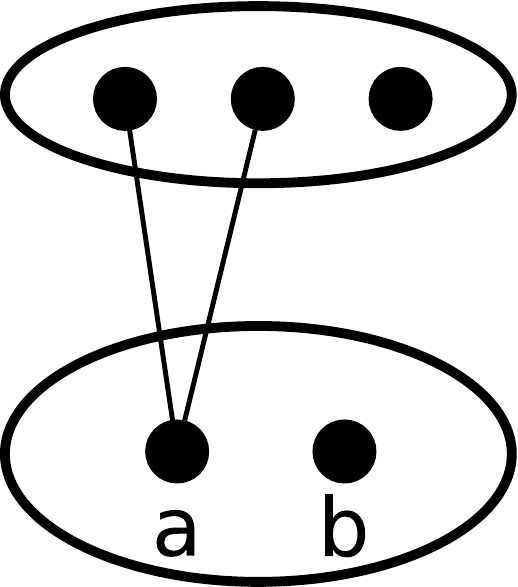}}
  \caption{\small ConNI: $a$ and $b$ are ConNI but not GNSub.}
  \label{CNI}
  \end{minipage}
  \end{figure}

\begin{theorem} {\em NSub and  GNSub are not comparable.}\end{theorem}
 Figures~\ref{fig:GNS} and~\ref{NSUB2} show counter examples.
\begin{theorem}{\em CtxDepI $\rightarrow$ GNSub}
\label{thm:CtxDepI-GNSub}
\end{theorem}
If $a$ and $b$ are CtxDepI, then the variable-value pairs connected to
$a$ in the micro-structure of the CSP are also connected to $b$.
Thus, $a$ and $b$ share at least one support and are GNSub.
Figure~\ref{GNS2} shows an example where the converse does not
hold.\footnote{Section~6.2 of \cite{zhang2004conditional} incorrectly
  states that CtxDepI and ConI are equivalent.}
\begin{theorem} {\em GNSub $\rightarrow$ ConNI.}
\label{thm:GNSub-ConNI}
\end{theorem}
If $a$ and $b$ are GNSub for a variable, then we can construct a set
of constraints that eliminates all values in the neighboring variables
except the ones that are the shared support between $a$ and $b$ in
order to make $a$ and $b$ ConNI. In Figure~\ref{CNI}, the constraints
that eliminate the supports of $a$ make $a$ and $b$ ConNI. 

\section{Relationships Between Interchangeability Concepts} \label{classification}

\begin{figure*} [ht]
\centering
\includegraphics[scale=0.35]{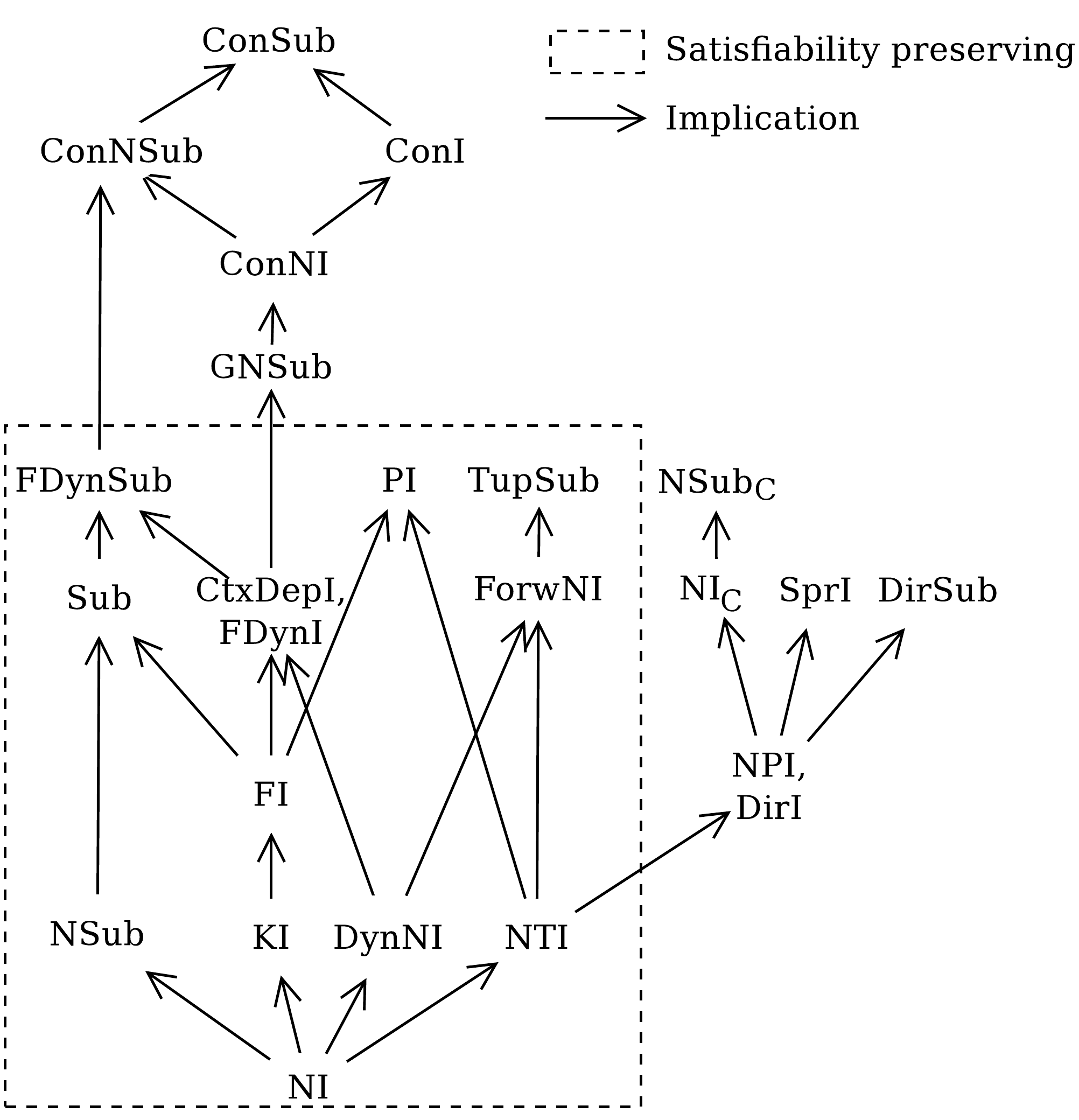}
\caption{\small Hasse Diagram of main interchangeability properties.}
\label{diagram}
\end{figure*}

\begin{figure*} [!ht]
\centering
\centerline{\includegraphics[viewport=0in .0in 9.8in
  5.8in,clip=true,scale=0.4]{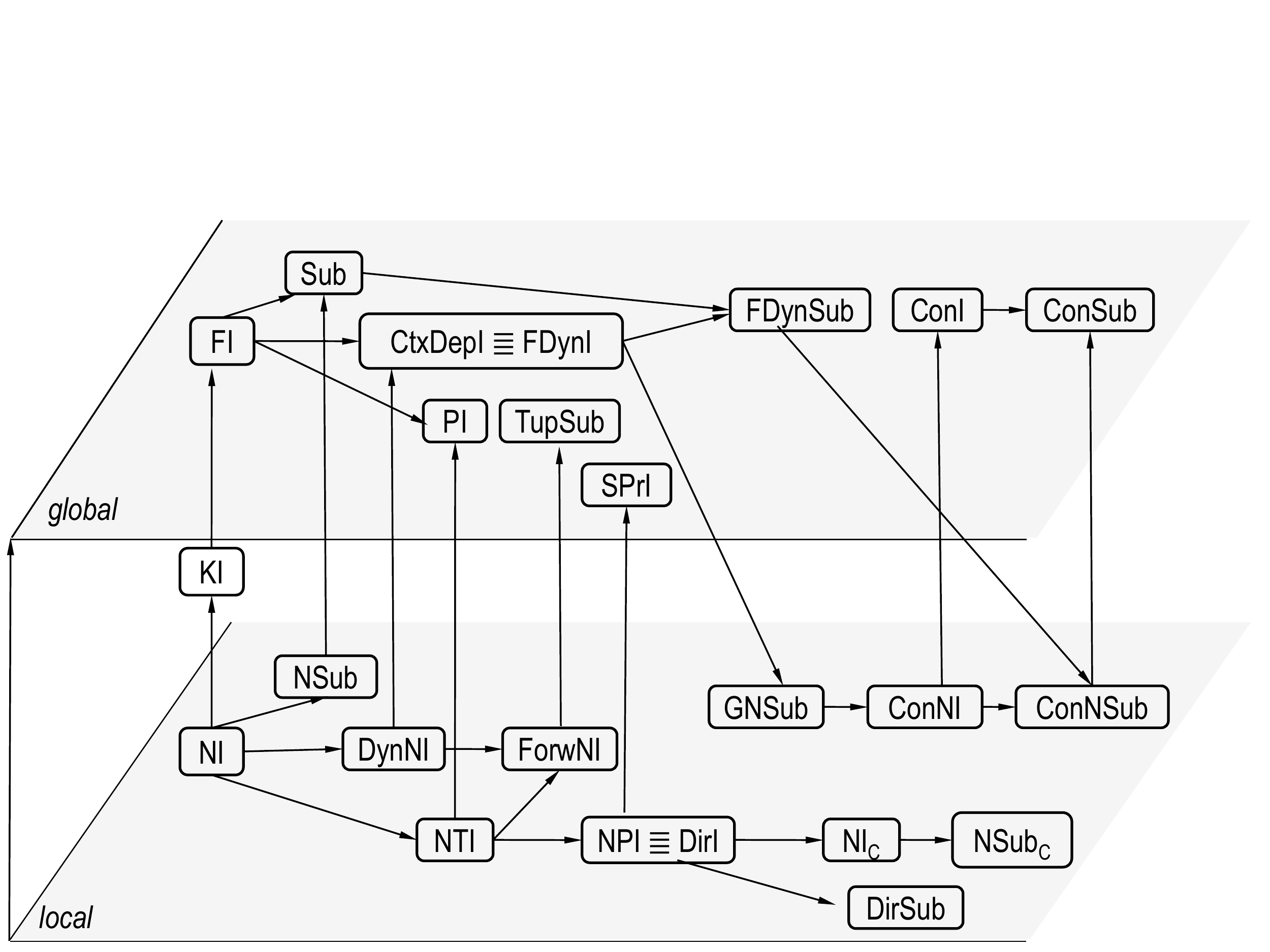}}
\caption{\small Depicting qualitative relations between main interchangeability properties.}
\label{diagram-h}
\end{figure*}

The different interchangeability concepts surveyed in the previous
sections are related here by the implication relation. Given two
interchangeability concepts $A$ and $B$, $A \rightarrow B$ if every
interchangeable pair defined by $A$, is also defined by $B$. Hence,
$B$ generalizes $A$. Figures~\ref{diagram} and~\ref{diagram-h}
illustrate those implication relations, and depict a partial ordering
because some concepts are not comparable.

An interchangeability concept is \emph{satisfiability preserving} iff
when given two values $a$ and $b$ that are either interchangeable or
$a$ is substitutable for $b$, removing $b$ from the problem does not
alter the satisfiability of the problem. Not all interchangeability
concepts are satisfiability preserving. Only the interchangeability
concepts that are inside the dashed rectangle are satisfiability
preserving.

In Figure~\ref{diagram-h}, The upper horizontal plane groups concepts
defined at the semantic level and thus are likely intractable.  The
lower horizontal plane groups concepts defined at the syntactic levels
(i.e., directly on the constraints), and can likely be efficiently
computed.

Interestingly, for a given form of interchangeability, when one moves
vertically upward from the lower plane to the higher plane,
interchangeability sets do not decrease in size while the
interchangeability form is not approximated or compromised.
Naturally, this advantage is not free because the computational cost
does not decrease.  Moving along the directed edges in either
horizontal planes does not increase cost or the opportunities for
interchangeability (i.e., size of interchangeability sets), but
results in an approximation (i.e., weakening) of the `quality' of the
interchangeability.  Finally, moving from the higher plane to the
lower one allows one to likely avoid intractability and may increase
the interchangeability opportunities but also results in
approximations that may lose solutions.

\section{Work in Progress} \label{conclusion}

In this paper we surveyed several forms of interchangeability,
presented their definitions, and analyzed their relationship.  This
document is a work in progress. In the expanded version, we will
address in depth the following topics:
\begin{description}
\item[More about interchangeability:] concepts not mentioned here;
missing proofs of incomparability; the satisfiability-preserving
property; restrictions to constraint types; algorithms for
interchangeability and their complexity; experimental results.

\item[Beyond classical CSPs:] soft constraints
\cite{cooper03,bistarelli-interchangeability2003,neagu2003computation,neagu2003soft}; 
distributed CSPs
\cite{petcu2004applying,Burke06applyinginterchangeability,ezzahir2007compilation}.

\item[Relation to symmetry:] various types
\cite{benhamou1994study,cohen2006symmetry};  symmetry breaking during search (SBDS)
\cite{roney2004tractable,BacWil02,Gent00symmetrybreaking}, symmetry
breaking by dominance detection (SBDD) \cite{fahetc01,focacci01}, and
symmetry breaking by enforcing variable or domain ordering
\cite{bellicha1994csp,yip2009evaluation}; restricted classes of
symmetries and problems where the symmetry is broken in polynomial
time \cite{van2003tractable,benhamou2004symmetry}.

\item[Relation to search-space compaction:] as in CPR \cite{HubbeF92},
  AND/OR graphs \cite{Dechter04theimpact}, SLDD
  \cite{wilson2005decision}, and solution robustness
  \cite{ginsberg98,hebrard-super}.

\item[Relation to SAT solving.]

\end{description}

\subsection*{Acknowledgments}

This work was supported in part by Science Foundation Ireland under
Grant 00/PI.1/C075.  Karakashian and Woodward gratefully acknowledge
the support and hospitality of the Cork Constraint Computation Centre
during Summer 2010 when this research was conducted.

\bibliography{biblio}
\bibliographystyle{named}

\end{document}